\documentclass[10pt,twocolumn,letterpaper]{article}

\usepackage[pagenumbers]{cvpr} 

\usepackage{booktabs}

\usepackage{amssymb}
\usepackage{pifont}

\newcommand{\cmark}{\ding{51}}
\newcommand{\xmark}{\ding{55}}

\newcommand{\agentname}{Vinedresser3D}

\usepackage{float} 

\definecolor{cvprblue}{rgb}{0.21,0.49,0.74}
\usepackage[pagebackref,breaklinks,colorlinks,allcolors=cvprblue]{hyperref}

\def\paperID{1527} 
\def\confName{CVPR}
\def\confYear{2026}

\newcommand{\equalcontrib}{\thanks{Equal contribution.}}
\newcommand{\samethanks}[1][\value{footnote}]{\footnotemark[#1]}

\title{\agentname: Agentic Text-guided 3D Editing} 

\author{
  Yankuan Chi$^{1}$\equalcontrib\hspace{0.23in}
  Xiang Li$^{2}$\samethanks\hspace{0.23in}
  Zixuan Huang$^{2}$\hspace{0.23in}
  James M. Rehg$^{2}$\\[1ex]
  $^{1}$The Hong Kong University of Science and Technology\\
  $^{2}$University of Illinois Urbana-Champaign\\
}

\begin{document}

\twocolumn[{%
\renewcommand\twocolumn[1][]{#1}%
\maketitle
\vspace{-7ex}
\begin{center}
    \centering
    \captionsetup{type=figure}
    \includegraphics[width=\textwidth,trim={0cm 0cm 0cm 0cm}, clip]{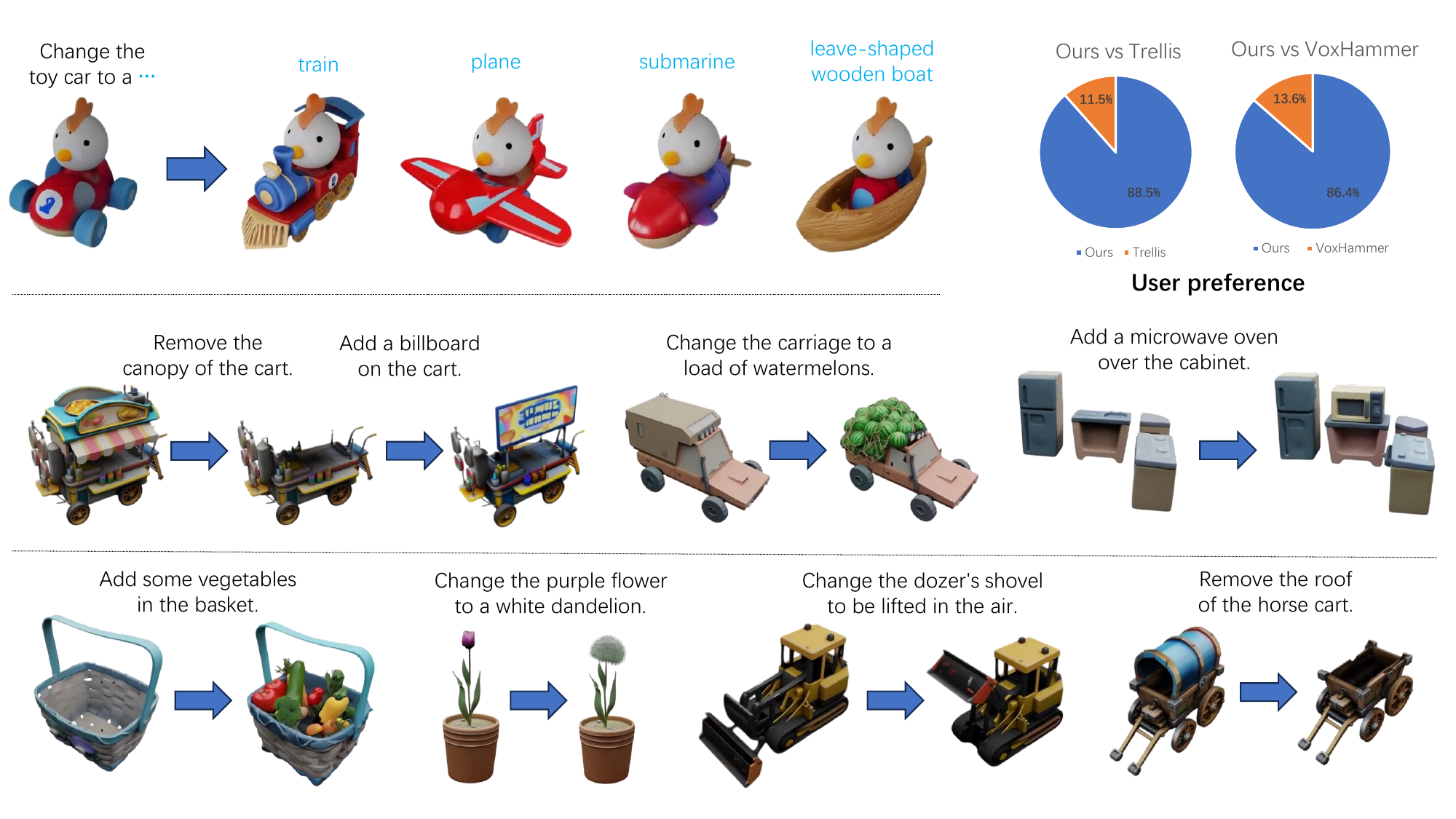}
    \vspace{-4ex}
    \captionof{figure}{We propose \agentname~, an agent that can intelligently perform high-quality text-guided 3D editing. It can handle various kinds of edits (addition, modification and deletion), support multi-turn editing and tackle different types of 3D assets (objects and scenes).}
\label{fig:teaser}
\end{center}%
\vspace{2ex}
}]

\begin{abstract}


Text-guided 3D editing aims to modify existing 3D assets using natural-language instructions. Current methods struggle to jointly understand complex prompts, automatically localize edits in 3D, and preserve unedited content. We introduce Vinedresser3D, an agentic framework for high-quality text-guided 3D editing that operates directly in the latent space of a native 3D generative model. Given a 3D asset and an editing prompt, Vinedresser3D uses a multimodal large language model to infer rich descriptions of the original asset, identify the edit region and edit type (addition, modification, deletion), and generate decomposed structural and appearance-level text guidance. The agent then selects an informative view and applies an image editing model to obtain visual guidance. Finally, an inversion-based rectified-flow inpainting pipeline with an interleaved sampling module performs editing in the 3D latent space, enforcing prompt alignment while maintaining 3D coherence and unedited regions. Experiments on diverse 3D edits demonstrate that Vinedresser3D outperforms prior baselines in both automatic metrics and human preference studies, while enabling precise, coherent, and mask-free 3D editing. 


\end{abstract}


\section{Introduction} \label{sec:introduction}

Editing 3D assets is a fundamental problem in 3D computer vision, with broad applications in digital content creation, virtual and augmented reality, and robotics. While recent text-guided 3D generation methods greatly lower the barrier to creating 3D content from scratch, in practice, high-quality 3D editing still relies heavily on professional artists and manual tools. This process is labor-intensive, requires domain expertise, and scales poorly with the growing demand for customized 3D content. These limitations motivate automatic systems that can follow high-level textual user instructions and perform precise 3D edits in an intelligent and reliable way.

Despite this great progress in 3D editing research, current text-guided 3D editing methods still struggle to semantically understand complex editing requests, automatically detect precise 3D editing regions from textual instructions alone, and perform high-quality 3D editing that both follows prompts closely and preserves unedited regions. Existing text-guided 3D editing methods can be broadly grouped into three lines, each with important limitations. Score Distillation Sampling-based approaches~\cite{Shape-Editor, Vox-E, Tip-Editor, 3DPaintBrush, Interactive3d, FocalDreamer, Gsedit, Plasticine3D, GaussianEditor} optimize a 3D representation under the guidance of 2D diffusion models, but per-scene optimization is computationally expensive and typically requires careful tuning to avoid unintended global changes. A second line follows a ``2D editing + 3D reconstruction'' paradigm~\cite{Tailor3d, MVEdit, Instant3dit, 3DEgo, VcEdit, Preditor3d, Phidias, MeshMaskLRM}. These works first edit rendered views with multi-view diffusion and image editing models and then reconstructing the edited asset with 3D reconstruction models. These methods are fundamentally constrained by multi-view inconsistency and incomplete 3D supervision, often struggling with the quality of unobserved regions and preservation of unedited geometry. Concurrent to our work, VoxHammer~\cite{VoxHammer} builds on a native 3D generative model~\cite{Trellis} to perform editing directly in 3D, but still requiring user-provided 3D masks, and cannot follow complex edit requests.

Recent advances in multimodal large language models (MLLMs)~\cite{geminiFlash}, image editing models~\cite{NanoBanana}, 3D segmentation~\cite{PartField}, and strong 3D flow-based native generative models~\cite{Trellis} have significantly improved our ability to understand text instructions, manipulate images, and operate directly in 3D latent space rather than through 2D proxies. Building on these developments, we argue that a natural next step is to move from \emph{single-model} methods to a \emph{3D editing agent} that integrates multiple specialized tools. Such an agent should (1) understand high-level text instructions, (2) automatically localize the intended editing region in 3D, and (3) invoke appropriate tools to carry out the edit while preserving the geometry and appearance of the rest of the asset.


We propose \agentname, an agent for high-quality text-guided 3D editing~(\Cref{fig:teaser}). \agentname~is built around a core of MLLM (Gemini-2.5-flash~\cite{geminiFlash}). It takes as input a 3D asset and a textual editing prompt. It first uses the MLLM to parse the instruction, reason about the desired changes, and produce detailed textual guidance for the edit. Then our agent selects a representative view and invokes an image editing model to obtain an edited image as visual guidance for 3D editing. To ensure the 3D editing does not make unintended local changes, our agent employs a 3D grounding pipeline that automatically detects the editing region in the 3D asset based on the text prompt, eliminating the need for user-provided 3D masks. Finally, leveraging the latent space of native 3D generation models such as Trellis~\cite{Trellis}, \agentname~performs precise inversion–based editing in 3D latent space, enabling faithful editing prompt alignment while preserving the geometry and appearance of unedited regions, producing a high-quality and holistic 3D edit.

In summary, our main contributions are:
\begin{itemize}
  \item We introduce \agentname, a 3D editing agent that uses an MLLM as its core to intelligently interpret text instructions and coordinate a set of tools for high-quality 3D editing. Our framework achieves strong text alignment, robust preservation of unedited parts, and high overall 3D quality. 
  \item We demonstrate that an MLLM trained primarily on 2D image--text data can be integrated into a 3D editing pipeline, where it plans editing strategies, generates multi-modal guidance, and interacts with 3D segmentation, image editing, and 3D generation tools, to greatly improve the editing performance. 
  \item We conduct extensive experiments on text-guided 3D editing, showing that \agentname~provides precise, coherent edits and compares favorably against state-of-the-art 3D editing baselines in both quantitative metrics and human evaluations.
\end{itemize}

\begin{figure*}[t]
\vspace{-1ex}
  \centering
   \resizebox{\textwidth}{!}{
   \includegraphics[width=\textwidth,trim={0cm 2.5cm 0cm 3cm}, clip]{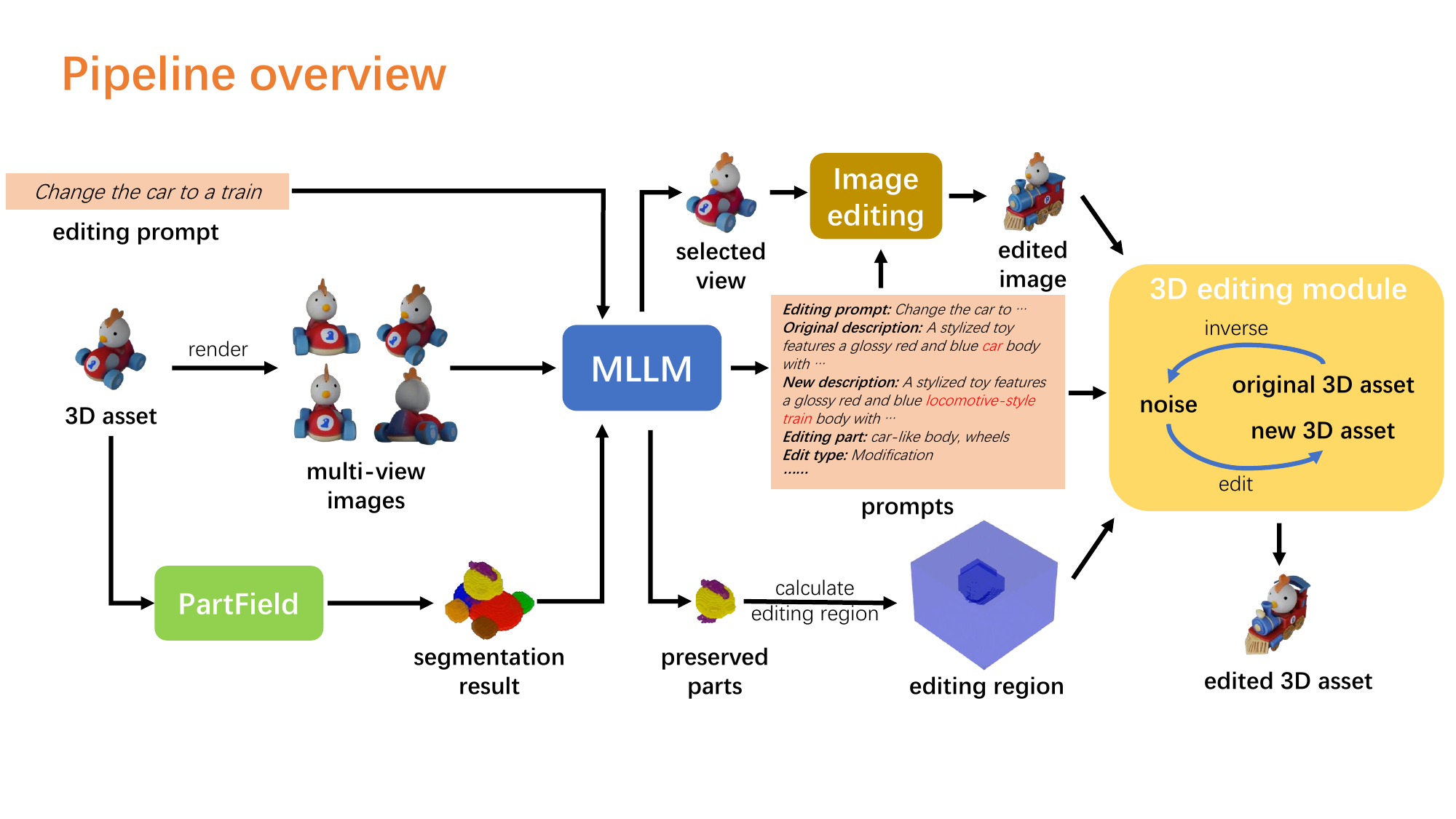}
   }
   \vspace{-4ex}
   \caption{Pipeline overview. Given a 3D asset and an editing prompt, \agentname~uses an MLLM to obtain new text and image guidance, automatically detects the intended editing region and then performs precise editing through an inversion-editing module.}
   \label{fig:pipeline}
   \vspace{-1ex}
\end{figure*}

\section{Related Work} \label{sec:RelatedWork}

\noindent\textbf{LLM Agents.} LLM Agents are of great research interest in the AI community in recent years~\cite{agent1, agent2, agent3, agent4, agent5, agent6, agent7}, since having an agent to automatically and smartly handle users' requests would be very convenient and efficient. Moreover, agents can utilize the strong knowledge learned by large AI models to solve users' problems accurately and effectively. In 3D generation and editing fields, many previous works~\cite{3DGeneAgent1, 3DGeneAgent2, 3DGeneAgent3, 3DGeneAgent4, 3DGeneAgent5, 3DGeneAgent6, 3DGeneAgent7, 3DGeneAgent8, 3DEditAgent1, 3DEditAgent2, 3DEditAgent3, 3DEditAgent4, 3DEditAgent5} have tried to develop intelligent agents to tackle the 3D tasks. However, no powerful agent for 3D editing with a text prompt as the pure input exists yet. So we believe developing an agent to intelligently perform text-guided 3D editing is of great importance. 

\noindent\textbf{Inversion-based editing.} Due to the limited amount of paired training data of editing, many editing pipelines~\cite{RF-Solver, InversionEditing1, InversionEditing2, InversionEditing3, InversionEditing4, InversionEditing5, InversionEditing6, InversionEditing7, InversionEditing8} adopt the inversion-editing methodology. It inverses the original object into the initial structured noise using a diffusion or flow model. It then edits the object following the generation process with new conditions towards the editing target. The editing process can either inject the features or attention maps of the inversion trajectory~\cite{RF-Solver, InversionEditing4, InversionEditing7} or inpaint the object~\cite{repaint}. We choose to use inpainting as we believe it can provide a greater control over parts we want to alter or preserve. Moreover, most inversion-based methods are for 2D editing because there are few powerful 3D diffusion or flow models. However, we want to directly operate in the 3D space, so we choose a start-of-the-art flow-based 3D generation model~\cite{Trellis} as our baseline.

\noindent\textbf{3D editing.} Due to the challenges of manipulating the 3D space and the lack of 3D data, most existing 3D editing methods actually heavily rely on the 2D domain. One stream of methods~\cite{Shape-Editor, Vox-E, Tip-Editor, 3DPaintBrush, Interactive3d, FocalDreamer, Gsedit, Plasticine3D, GaussianEditor} utilizing Score Distillation Sampling~\cite{DreamFusion} to update the 3D asset. They repeatedly render images and propagate the gradients from a 2D diffusion model back to the 3D asset to optimize it. However, this process is time-consuming and computationally expensive. Another stream of methods~\cite{Tailor3d, MVEdit, Instant3dit, 3DEgo, VcEdit, Preditor3d, Phidias, MeshMaskLRM} render multi-view images of the original assets and edit the images with multi-view diffusion models. They then reconstruct the edited asset from those images using reconstruction techniques. This stream of methods usually suffers from multi-view inconsistency and the lose of spatial information due to occlusion and distortion of rendering. Considering the problems with relying on the 2D domain, some methods~\cite{Instructp2p, ShapeTalk, ShapeLLM-Omni, Kyvo} propose to encode the 3D asset into a latent space and perform editing on it. But due to the limitation of available paired 3D editing training data, they can not achieve satisfactory results. Recent flow-based native 3D generation models~\cite{Trellis, Triposg} provide the potential of directly editing in the 3D space. And we decide to choose one of them~\cite{Trellis} as our baseline to avoid the problems of heavily relying on the 2D domain and inadequate paired 3D editing data.
\section{Method} \label{sec:method}

We first provide the preliminaries in \cref{sec:method_preliminary}. Given the original 3D asset and the editing prompt as the only input, \agentname~utilizes an MLLM (Gemini-2.5-flash~\cite{geminiFlash}) as its core to obtain new guidance (\cref{sec:method_guidance}) and detect the intended editing region (\cref{sec:method_region}). After that, \agentname~performs 3D editing to get the new asset through an inversion-based Trellis module (\cref{sec:method_interleavedTrellis}) with the guidance and the editing region. The pipeline overview is in \cref{fig:pipeline}.

\subsection{Preliminary} \label{sec:method_preliminary}

\noindent\textbf{Trellis}~\cite{Trellis} is a flow-based native 3D generation model. It takes either a text prompt or a single image as the condition, and generate a corresponding 3D asset. Trellis employs a structured latent representation (SLAT), which represents a 3D asset as sparse voxels with their latent features. 
The generation pipeline is decomposed into two stages, each utilizing a specialized rectified flow model~\cite{rectifiedFlow}. The first stage predicts the sparse voxel structure, while the second generates the detailed latent features for each voxel. The resulting SLAT can be  decoded into 3D Gaussians~\cite{3DGS} or meshes. We adopt Trellis as our baseline given its state-of-the-art performance in high-fidelity 3D generation.

\begin{figure}[t]
  \centering
   \resizebox{0.49\textwidth}{!}{
   \includegraphics[width=\textwidth,trim={0cm 0cm 16cm 0cm}, clip]{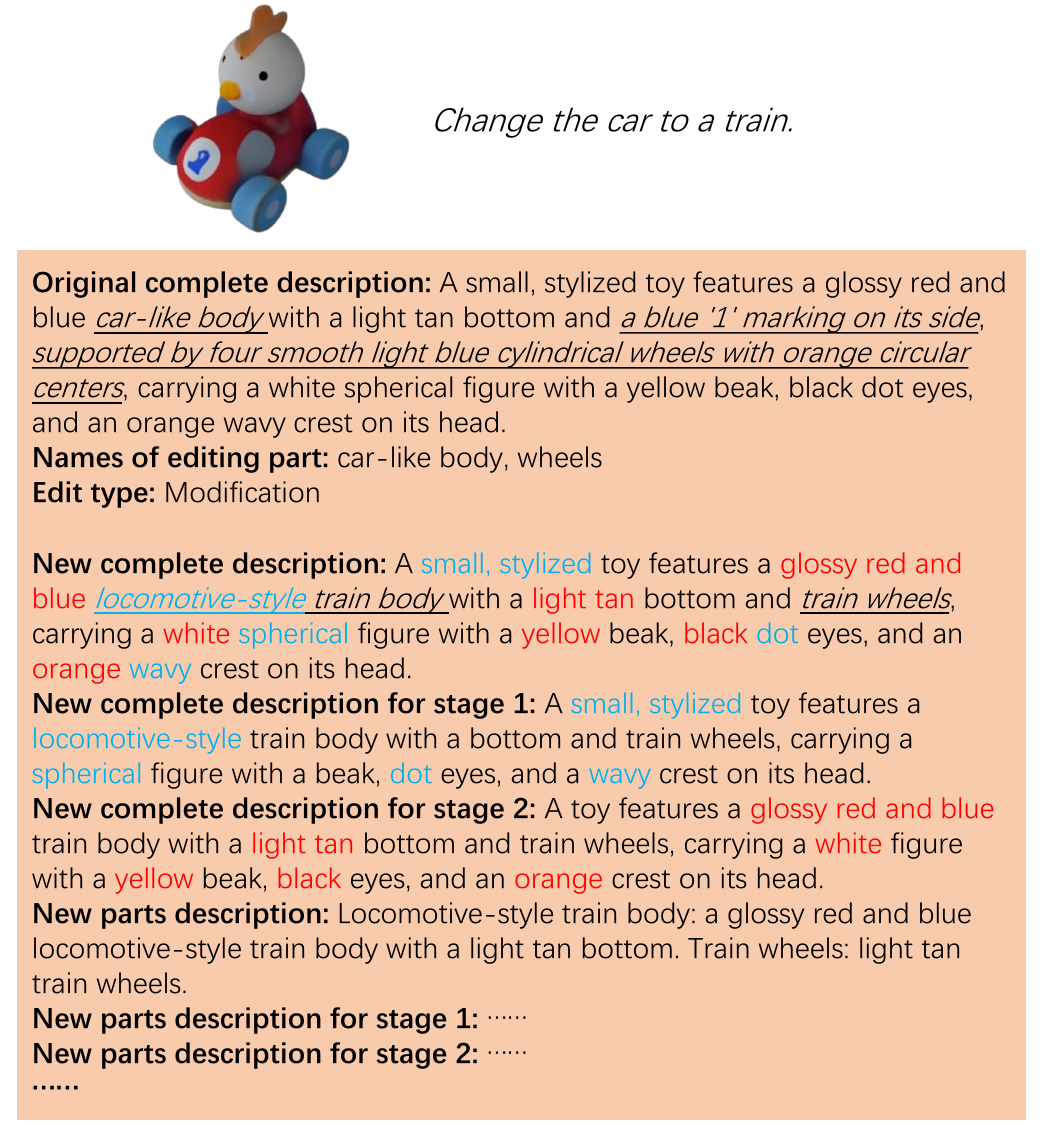}
   }
   \vspace{-3ex}
   \caption{Text guidance output by the MLLM. The modified words between the original complete description and the new complete description are marked with underlined italics. We highlight the extracted stage 1-related (in \textcolor{cyan}{cyan}) and stage 2-related (in \textcolor{red}{red}) information.}
   \label{fig:txtGuide}
   \vspace{-1ex}
\end{figure}


\noindent\textbf{RF-Solver}~\cite{RF-Solver} introduces a method for the accurate inversion of rectified flow models in the context of image generation. While standard inversion typically relies on first-order discretization:
\begin{equation}
  X_{i-1} = X_i + (t_{i-1} - t_i)v_{\theta}(X_i,t_i),
  \label{eq:vanilla_inv}
\end{equation}
where $X_i$ represents the noisy state at timestep $t_i$ and $v_{\theta}$ denotes the flow velocity field, RF-Solver enhances this by incorporating a second-order Taylor expansion term:
\begin{equation}
    X_{i-1} = X_i + (t_{i-1} - t_i)v_{\theta}(X_i,t_i) + \frac{1}{2}(t_{i-1}-t_i)^2 v_{\theta}^{(1)}(X_i,t_i).
    \label{eq:RF-solver}
\end{equation}
Here, $v_{\theta}^{(1)}$ represents the temporal derivative of the velocity field. We adopt RF-Solver in our 3D editing module because this second-order term significantly improves inversion fidelity. 

\begin{figure*}[t]
\vspace{-1ex}
  \centering
   \resizebox{0.8\textwidth}{!}{
   \includegraphics[width=\textwidth,trim={3cm 0.8cm 2cm 0cm}, clip]{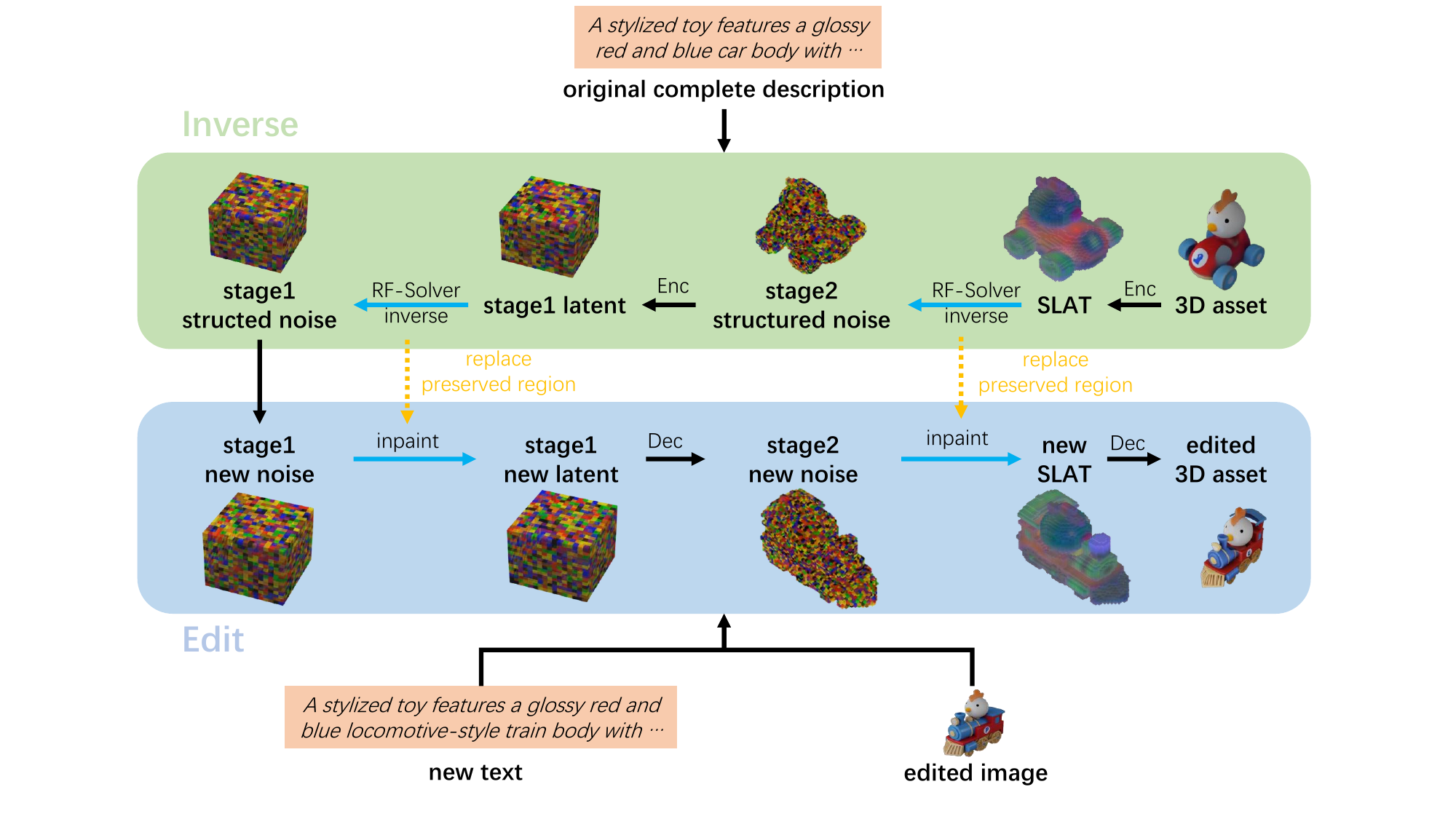}
   }
   \caption{Our native 3D inversion-based editing pipeline. It first invert the original 3D asset back to structured noises using RF-Solver~\cite{RF-Solver} and the original complete description as the condition. Then it performs editing through inpainting by denoising with Trellis-text and Trellis-image alternatively for all timesteps, using both the new text and edited image as conditions.}
   \label{fig:interTre}
\end{figure*}


\subsection{Multi-modal Guidance Generation} \label{sec:method_guidance}
Given the input 3D asset and a user-provided editing prompt, \agentname~first generates the necessary text and image guidance to condition downstream modules. Leveraging the reasoning capabilities of a Multimodal Large Language Model (MLLM), the agent semantically understand the editing request to generate guidance signals that are both comprehensive and precisely aligned with the target modification.


\agentname~first generates the text guidance (\cref{fig:txtGuide}). The process begins with rendering multi-view images of the original 3D asset. These images, combined with the user's editing prompt, are provided as input to the MLLM. We employ a multi-step prompting strategy: first, the MLLM analyzes the input to output a description of the original asset, identify the specific names of parts targeted for editing, and classify the editing operation type (addition, modification, or deletion). Then, the model is prompted to predict the complete description of the asset post-editing. Importantly, we constrain the MLLM to maximize semantic preservation of the original description outside the targeted edit regions to retrain fine-grained details. Next, the MLLM isolates descriptions specifically for the newly modified target parts. Finally, to align with our two-stage generation pipeline, the MLLM decomposes these descriptions into structure-related components (for stage 1 geometry) and appearance-related components (for stage 2 latent features). This multi-stage process provides highly targeted conditioning for subsequent modules. As shown in \cref{fig:txtGuide}, the MLLM demonstrates implicit 3D semantic understanding despite its 2D training, accurately processing spatial information.


Furthermore, \agentname~generates image guidance through a view-selection and editing pipeline. The MLLM evaluates the rendered views to identify the candidate that maximizes the visibility of the target editing parts and the overall asset structure. The selected image is then forwarded to an image editing model (Nano Banana~\cite{NanoBanana}). We augment the conditioning of the image editing model by supplying both the editing prompt and the decoupled description of the new target parts. This multi-level text conditioning guides the generation of a high-fidelity reference image that accurately reflects the desired edits.

\subsection{Detect editing region} \label{sec:method_region}

One main advantage of \agentname~over previous work is that it can automatically detect the editing region, so users do not need to provide a 3D mask.
For modification and deletion requests, \agentname~first identifies which parts of the original asset $A$ should be edited and which should be preserved, denoted by $P_{\text{edit}}$ and $P_{\text{pres}}$, respectively. To this end, it employs PartField~\cite{PartField}, a powerful 3D segmentation model, to decompose $A$ into $S$ semantic parts. The original 3D asset and its segmented colored view (rendered as multi-view images), together with text guidance specifying the target parts, are then fed into the MLLM, which selects the editing region $P_{\text{edit}}$ and implicitly defines the preserved region as $P_{\text{pres}} = A \setminus P_{\text{edit}}$. Since PartField’s segmentation granularity depends on $S$, \agentname~computes segmentations for multiple values of $S$ and passes all of them to the MLLM, allowing it to choose the most fine-grained and semantically precise parts.

After partitioning the original asset $A$ into $P_{\text{edit}}$ and $P_{\text{pres}}$, \agentname~must determine the corresponding editing region $R_{\text{edit}}$ (and its complement, the preserved region $R_{\text{pres}} = C \backslash R_{\text{edit}}$, where $C$ denotes the entire voxel grid). This is straightforward for addition and deletion requests: for addition, $R_{\text{edit}}$ is all voxels in the 3D space outside the original asset ($C \backslash A$), and for deletion, $R_{\text{edit}}$ coincides with $P_{\text{edit}}$. In contrast, computing $R_{\text{edit}}$ for modification requests is non-trivial. We cannot simply treat all voxels outside $P_{\text{pres}}$ as editable, because Trellis may modify a layer of voxels above the preserved geometry, unintentionally altering $P_{\text{pres}}$.

To address this, we introduce the following procedure. We first assign all voxels in $P_{\text{edit}}$ to $R_{\text{edit}}$ and all voxels in $P_{\text{pres}}$ to $R_{\text{pres}}$. We then compute a bounding box for each preserved part and denote the union of these boxes as $bbox_{\text{pres}}$. All empty voxels outside $bbox_{\text{pres}}$ are also assigned to $R_{\text{edit}}$. Next, for each empty voxel $v$ inside $bbox_{\text{pres}}$, we find its $k$-nearest voxels belonging to the original asset and compute the fraction of those lying in $P_{\text{edit}}$, denoted as $\text{PropKNN}(v)$. If this proportion exceeds a threshold $\tau$, we classify $v$ as editable; otherwise, it is preserved. Formally, the editing region is defined as:
\begin{equation}
  R_{\text{edit}} = 
  \begin{cases}
      C \backslash A & \text{addition}\\
      P_{\text{edit}} & \text{deletion}\\
      P_{\text{edit}} \cup (C \backslash bbox_{\text{pres}}) \cup V & \text{modification}\\
  \end{cases} \\   
  \label{eq:compute_editing_region}
\end{equation}
$\text{where }V = \{v | v \in bbox_{\text{pres}} \backslash A,  \text{PropKNN}(v)>\tau \}$.

This automatic localizing pipeline equips our agent with some spatial reasoning ability. As we can see in \cref{fig:quali}, \agentname~ can leverage the world knowledge and common sense learnt by the MLLM to understand the user's intention and  precisely locate the desired editing region in a complex 3D asset. 


\begin{table*}[t]
  \centering
  \resizebox{0.7\textwidth}{!}{
  \begin{tabular}{l|c|c|cccc|c}
      \toprule
      & & Text Align. & \multicolumn{4}{|c|}{Unedited Preservation} & 3D Quality \\
        \midrule
      Method           & Human Mask & CLIP-T$\uparrow$  & CD$\downarrow$ & PSNR$\uparrow$ & SSIM$\uparrow$ & LPIPS$\downarrow$ & FID$\downarrow$  \\
        \midrule
    Instant3dit~\cite{Instant3dit} & \cmark & 0.227 & 0.027 & 20.86 & 0.851 & 0.153 & 80.35  \\
    VoxHammer~\cite{VoxHammer} & \cmark & 0.235 & 0.027 & 24.36 & 0.890 & 0.087 & 34.95  \\
    Trellis~\cite{Trellis} & \cmark & 0.247 & 0.010 & 37.35 & \textbf{0.984} & 0.017 & 31.10   \\
    Ours & \xmark & \textbf{0.252}  &  0.016 & 29.45 & 0.953 & 0.045 & 29.49 \\
    Ours w/ HM  & \cmark & \textbf{0.252} &  \textbf{0.008} & \textbf{37.69} & \textbf{0.984} & \textbf{0.015} & \textbf{27.38}  \\
    \bottomrule
    \end{tabular}
   }
    \caption{Quantitative comparison. We include the results of our method with human-provided 3D masks (Ours w/ HM). Best results are in \textbf{bold}. Our method with human-provided 3D masks achieves the best results in all metrics. Even though our method without human-provided 3D masks does not preserve the unedited parts as well as Trellis, it still closely aligns with the editing prompt and produces high-quality outcomes.
  }
  \label{tab:main_quan}
\end{table*}

\begin{table}[t]
  \centering
  \resizebox{0.45\textwidth}{!}{
  \begin{tabular}{l|ccc}
  \toprule
 Model & Text Align. & Unedited Preservation & 3D Quality \\
 \midrule
 vs. Trellis & 92.5\% & 82.0\% & 90.8\% \\
 vs. VoxHammer & 89.8\% & 79.3\% & 90.2\% \\
 \bottomrule
  \end{tabular}}
 \caption{User study. We ask the user to select the better one between ours and another method in terms of editing prompt alignment, unedited parts preservation and overall 3D quality. We report the win rates of our methods. Our method achieves high win rates in all perspectives.}
  \label{tab:user_study}
\end{table}


\subsection{Inversion-Based 3D Editing} \label{sec:method_interleavedTrellis}
Given the multi-modal guidance and the 3D mask $R_{\text{edit}}$, \agentname~performs localized 3D editing via an inversion-inpainting process. The asset is first inverted to its latent initial noise representation, and then edited through inpainting. We implement this through a novel inversion-based module (\cref{fig:interTre}) that combines Trellis-text and Trellis-image models to achieve both high-fidelity detail and comprehensive structural coherence in the edited results.



We employ RF-Solver~\cite{RF-Solver} for accurate inversion in the 3D flow models. Conditioned on the complete description of the original asset, the process begins with Stage 2 inversion to derive the structured noise for the SLATs. Then, the sparse voxel structure is encoded into the Stage 1 latent space for Stage 1 inversion, yielding the corresponding noise. Empirically, we observed through grid search that setting the Classifier-Free Guidance (CFG)~\cite{CFG} strength to 0 significantly stabilizes the inversion trajectory and minimizes reconstruction error. Therefore, we use a CFG scale of 0 for all inversion steps. Given the inverted noises, we implement a mask-guided inpainting strategy to synthesize the edited asset while preserving the identity of unedited regions. Specifically, at each timestep of the denoising process, the latent features of voxels located outside the edit mask $R_{\text{edit}}$ are replaced with their spatially corresponding features from the original inversion trajectory. 
Throughout this generation process, the flow models are conditioned on the decomposed Stage 1 and Stage 2 text guidance, along with the generated image guidance.


We observe that directly applying inversion-based editing with either Trellis-text or Trellis-image alone is insufficient. 
Trellis-text produces limited generation quality, particularly in fine details, due to the scarcity of text-aligned 3D training data~\cite{Trellis}. Trellis-image is conditioned on a single viewpoint, and therefore struggles with occluded regions where visual information is unavailable. To address these limitations, we introduce an Interleaved Trellis editing Module that leverages the complementary strengths of both modalities. We construct the denoising trajectory by alternating between one timestep of Trellis-text denoising and one timestep of Trellis-image denoising, interleaving the vector fields of the two models. This design combines the broad semantic alignment and prompt adherence from the text branch with the high-fidelity detail from the image branch, leading to improved overall generation quality.

We now describe additional design details for mask handling across stages. Note that Trellis produces the final asset at $64^3$ voxel resolution, while Stage 1 operates in a $16^3$ latent space. We downscale the editing mask accordingly via proportion thresholding to obtain the Stage 1 mask. For Stage 2 editing, inspired by~\cite{VoxHammer}, we adopt a soft mask instead of a hard mask. Specifically, for each voxel in the preserved region, we compute its distance to the editing region. For voxels close to the boundary, we compute their features as a weighted average of the denoised and inverted features during editing, rather than directly replacing them with the inverted ones. We find that this soft masking effectively eliminates floating artifacts at the boundary of the preserved region. For deletion requests, we skip Stage 1 inversion and editing entirely. Instead, we directly remove all voxels in $R_{\text{edit}}$ and apply Stage 2 to smooth the boundary.
\section{Experiment} \label{sec:experiment}

\begin{figure*}[t]
  \centering
   \resizebox{\textwidth}{!}{
   \includegraphics[width=\textwidth,trim={0cm 15cm 0cm 0cm}, clip]{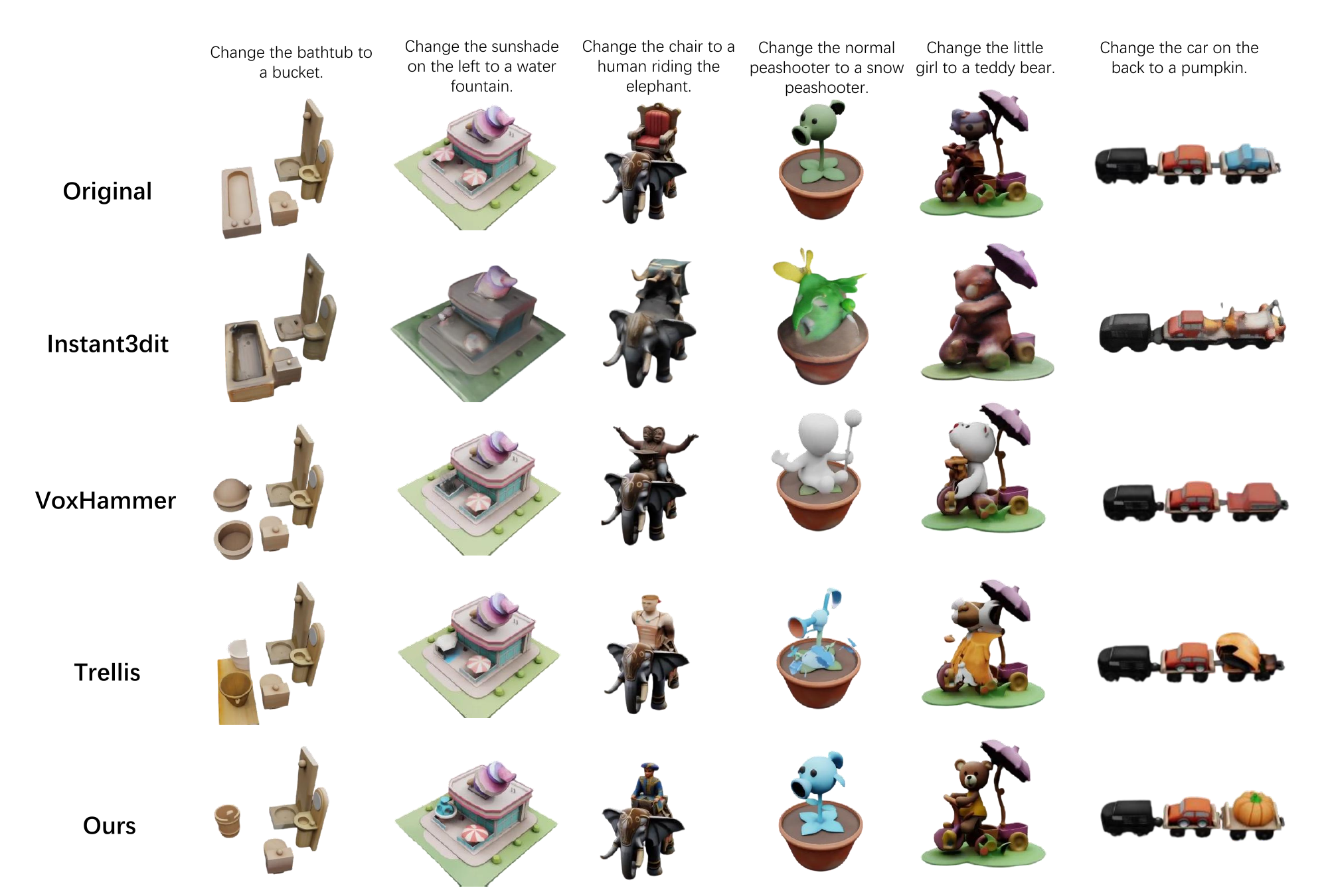}
   }
   \vspace{-1ex}
   \caption{Qualitative comparison of different methods. We can see that our method surpasses all the others by smartly interpreting the editing intention of the user, closely following the editing prompt, precisely locating the intended editing region and generating high-fidelity results.}
   \label{fig:quali}
\end{figure*}

\subsection{Setup}

\noindent\textbf{Implementation details.} We adopt Google Gemini-2.5-Flash~\cite{geminiFlash} as the MLLM and Nano Banana~\cite{NanoBanana} as the image editing model. We render 8 views of each asset, for both the original appearance and the segmentation, as input to the MLLM for text guidance and editing region selection. We additionally render 24 views from varying rotations and elevations for the MLLM to select the best view as image guidance. Since PartField~\cite{PartField} operates on point clouds for both input and output, we decode each asset into 3D Gaussians as point clouds before feeding them into PartField, and convert the resulting segmentation back to voxels via majority voting. The number of semantic parts is set to $S \in [3,8]$. We follow Trellis's~\cite{Trellis} configuration for the 3D editing module. During editing, the agent explores different combinations of positive and negative prompts as text conditions and selects the best result.

\noindent\textbf{Dataset.} We collect high-quality and diverse 3D assets from both model-generated and human-created sources. Specifically, we gather 24 assets from Trellis generation results and 33 assets from GSO~\cite{GSO} and PartObjaverse-Tiny~\cite{PartObjaverseTiny}. For each asset, we carefully design an editing prompt that aligns with common sense. Our prompts cover a wide range of editing categories and difficulty levels.


\noindent\textbf{Baselines.} We compare our method with three state-of-the-art 3D editing pipelines: Trellis~\cite{Trellis} editing, VoxHammer~\cite{VoxHammer}, and Instant3dit~\cite{Instant3dit}. Trellis supports 3D editing through human-provided 3D masks combined with Repaint~\cite{repaint}. VoxHammer builds on Trellis-image by rendering an image, applying image inpainting to obtain a new condition, and performing inversion-based 3D editing, but it still requires a human-provided 3D mask. Instant3dit is a multi-view diffusion and reconstruction-based editing model.


\noindent\textbf{Metrics.} We use CLIP-T~\cite{CLIPT} to measure alignment with the editing prompt, Chamfer Distance~\cite{CD}, PSNR, SSIM~\cite{SSIM}, and LPIPS~\cite{LPIPS} to assess preservation of unedited regions, and FID~\cite{FID} to evaluate overall 3D quality. We also conduct a user study to evaluate human preference.


\subsection{Qualitative Results}
We present qualitative comparisons in \cref{fig:quali}, where our method surpasses all baselines. It accurately interprets the user's editing intention and faithfully follows the editing prompt to produce expected outcomes, demonstrating the effectiveness of the MLLM and the guidance it generates. Additionally, our method precisely localizes the target editing region within complex assets given only text prompts, validating the editing region detection pipeline. The overall quality of the assets is also well preserved, confirming the competence of the inversion-based Trellis editing module.


\subsection{Quantitative Results}
We present quantitative comparisons in \cref{tab:main_quan}. Since all baselines require human-provided 3D masks, we also include results of our method with human-provided masks for fair comparison. Both variants of our method, with and without human-provided masks, achieve the best CLIP-T score. This validates the benefit of leveraging an MLLM to produce detailed text guidance. For unedited region preservation metrics (\ie CD, PSNR, SSIM, LPIPS), our method remains competitive even without human provided masks. Given human-provided masks, our approach achieves the best results. Finally, both settings of our method outperform all baselines in overall 3D quality (FID).


We also conduct a user study to evaluate human preference across methods. Users are asked to choose the better output between two results from three perspectives: editing prompt alignment, unedited region preservation, and overall 3D quality. The win rates are reported in \cref{tab:user_study}. The results clearly show that our method consistently outperforms both Trellis and VoxHammer across all perspectives.

\begin{figure}[t]
  \centering
   \resizebox{0.45\textwidth}{!}{
   \includegraphics[width=\textwidth,trim={3cm 2cm 6cm 1cm}, clip]{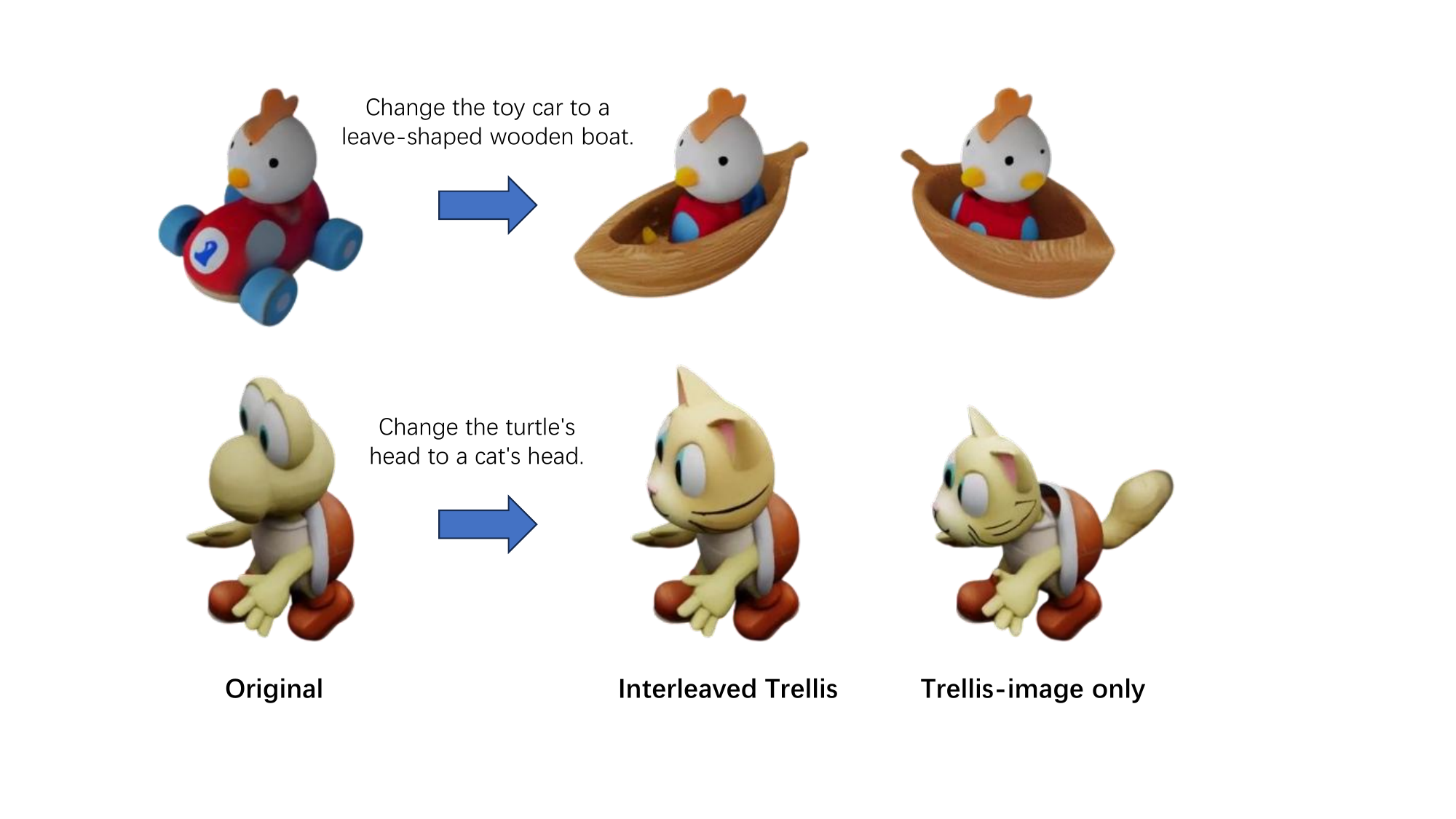}
   }
   \caption{Qualitative ablation of interleaved Trellis vs Trellis-image only. The outputs of Trellis-image only editing may be distorted or unreasonable.}
   \label{fig:ablaImg}
\end{figure}

\begin{table}[t]
  \centering
  \resizebox{0.45\textwidth}{!}{
  \begin{tabular}{l|ccc|c}
  \toprule
  & \multicolumn{3}{|c|}{Unedited Preservation} & 3D Quality \\
  \midrule
 Methods & PSNR$\uparrow$ & SSIM$\uparrow$ & LPIPS$\downarrow$ & FID$\downarrow$ \\
  \midrule
 Ours & \textbf{29.45} & \textbf{0.953} & \textbf{0.045} & \textbf{29.49} \\
 \midrule
 Ours w/o Trellis-text & 28.06 & 0.943 & 0.054 & 30.59 \\
 Ours w/o $R_{\text{edit}}$ & 25.65 & 0.921 & 0.068 & 33.95 \\
 \bottomrule
 \end{tabular}}
 \caption{Quantitative results of the ablation study. We compare our method with the Trellis-image only 3D editing design and ours without the detected editing region. We use PSNR, SSIM and LPIPS to measure unedited parts preservation and FID to measure the overall 3D quality. We can see that our method outperforms both of the ablated ones on all metrics.}
  \label{tab:abla}
\end{table}




\subsection{Ablation Study}
We conduct two ablation experiments to validate key components of our agent. The first ablation uses only Trellis-image in the editing module, verifying the effect of the interleaved Trellis design. The second ablation removes the editing region mask (\ie, performing purely interleaved Trellis generation instead of inpainting), verifying the necessity of the editing region detection module.

When using only Trellis-image, the editing module receives limited information due to occlusion and deformation during rendering, which can lead to distorted or unreasonable outputs. We provide qualitative examples in \cref{fig:ablaImg} and report FID scores in \cref{tab:abla}, demonstrating the importance of interleaving Trellis-text and Trellis-image for overall 3D quality.

\begin{figure}[t]
  \centering
   \resizebox{0.45\textwidth}{!}{
   \includegraphics[width=\textwidth,trim={3cm 2cm 6cm 1cm}, clip]{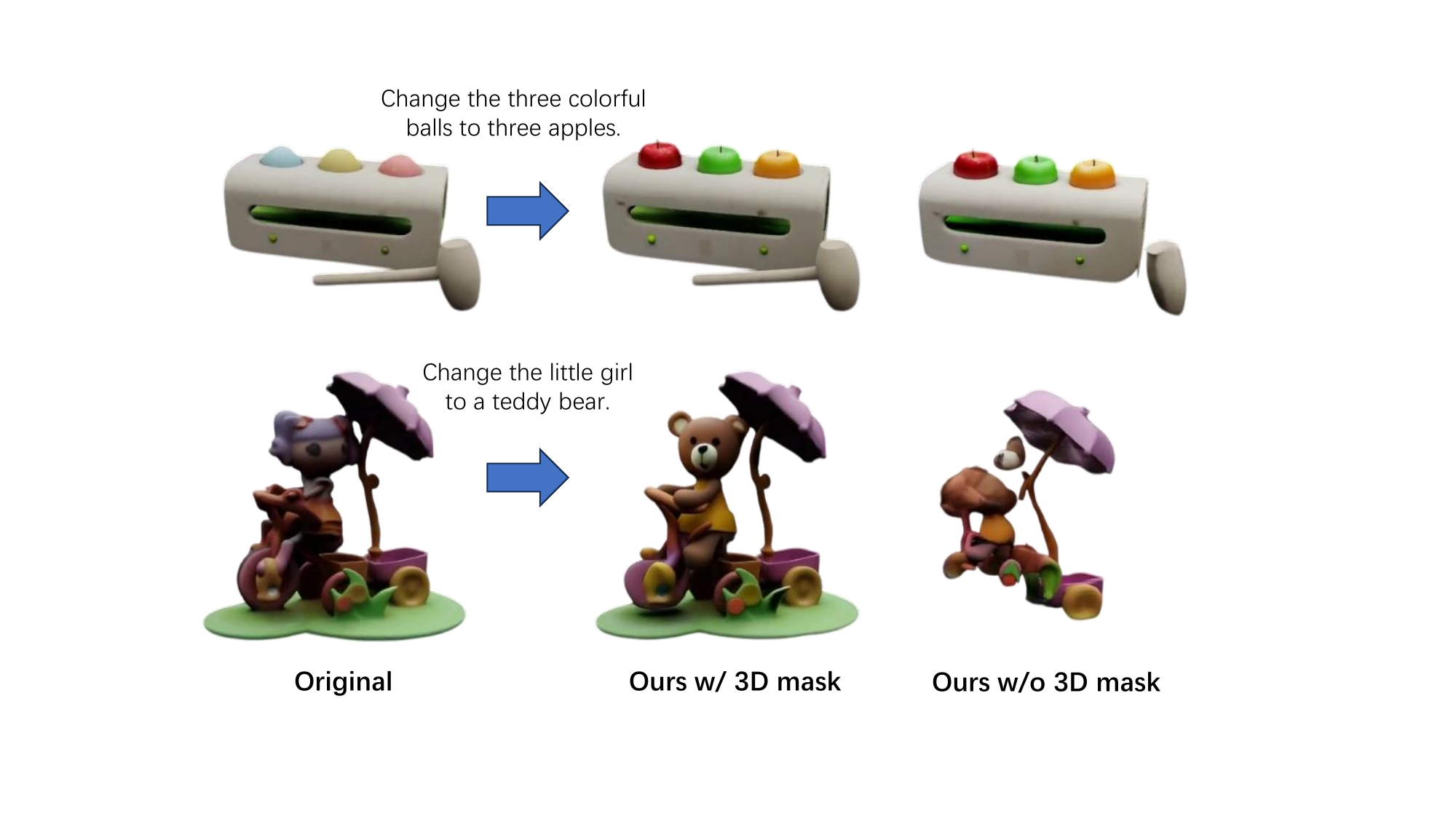}
   }
   \caption{Qualitative ablation of our method with and without the detected editing region as the 3D mask. Our method without the detected editing region may alter the intended preserved parts or produce distorted outputs.}
   \label{fig:ablaMask}
\end{figure}

In the second ablation, we verify that providing the editing region to the editing module helps preserve unedited parts (the first example in \cref{fig:ablaMask} and PSNR, SSIM, LPIPS in \cref{tab:abla}). It also plays an important role in maintaining the overall quality of the edited asset (the second example in \cref{fig:ablaMask} and FID in \cref{tab:abla}). Without a mask, the interleaved Trellis editing can fail to handle the asset properly and produce distorted outputs. Providing a 3D mask and injecting preserved-region features during editing regularizes the denoising process and helps maintain overall quality.


\section{Limitation} \label{sec:limitation}


While \agentname~can produce impressive results, it still has several limitations. First, the underlying MLLM does not accept native 3D input. We expect that enabling the MLLM to directly consume 3D inputs and perform 3D reasoning could further improve the results. Second, the external tools invoked by \agentname~are imperfect. For example, PartField~\cite{PartField} can produce unreasonable part segmentations. We anticipate that future advances in these 3D models will further improve the overall performance of our agent.
\section{Conclusion} \label{sec:conclusion}

We develop an agent that is capable of intelligently and efficiently performing high-quality text-guided 3D editing. Using an MLLM as its core, it smartly makes use of an image editing model, a 3D segmentation model and a 3D generation baseline. It intelligently interprets the editing request, generates detailed and comprehensive guidance, automatically detects the intended editing region and performs precise editing. It produce high-quality results in terms of editing prompt alignment, unedited parts preservation and overall 3D quality. Also, we show the potential of integrating 2D MLLMs into 3D pipelines. We believe our work will promote the 3D editing field towards an agentic, high-quality and smart future.
{
    \small
    \bibliographystyle{ieeenat_fullname}
    \bibliography{main}
}


\end{document}